\documentclass[letterpaper, 10 pt, conference]{ieeeconf}

\IEEEoverridecommandlockouts                              %
\usepackage[T1]{fontenc}
\usepackage[utf8]{inputenc}

\usepackage{amsmath} %
\usepackage{amssymb} %

\usepackage{hyperref}
\usepackage{graphicx}
\usepackage{tabularx}
\usepackage{xcolor}
\usepackage{colortbl}
\usepackage{array}
\newcolumntype{Y}{>{\raggedright\arraybackslash}X}
\definecolor{TopHeaderBg}{HTML}{575AB1} %
\definecolor{HeaderBg}{HTML}{2B3F9A}    %
\definecolor{HeaderFg}{HTML}{FFFFFF}
\definecolor{RowAlt}{HTML}{F6F6F6}
\usepackage{multirow}
\usepackage{booktabs}
\usepackage{adjustbox, tabularx, booktabs, array, makecell, xcolor, colortbl}
\definecolor{tableheader}{HTML}{EDEFFA}

\usepackage{cleveref}
\usepackage{cite}

\PassOptionsToPackage{dvipsnames,table}{xcolor}

\usepackage[frozencache,cachedir=.]{minted}
\setminted{
  linenos,
  breaklines=true,
  fontsize=\small,
  numbersep=8pt,
  xleftmargin=2em
}

\usepackage{booktabs, tabularx}
\usepackage[table]{xcolor}

\definecolor{HeaderBg}{HTML}{575AB1}  %
\definecolor{HeaderFg}{HTML}{FFFFFF}
\definecolor{RowAlt}{gray}{0.97}      %

\usepackage{tabularx,booktabs,xcolor,makecell}
\definecolor{tableheader}{HTML}{F2F2F7} %
\definecolor{groupsep}{HTML}{D9D9E6}    %

\usepackage{bm}
\usepackage{url}

\usepackage{ifthen}

\title{\LARGE \bf Event-LAB: Towards Standardized Evaluation of\\ Neuromorphic Localization Methods}

\author{Adam D Hines$^{\ast\space1,2}$ \quad Alejandro Fontan$^{1}$ \quad Michael Milford$^{1}$ \quad Tobias Fischer$^{1}$
\thanks{$^{1}$These authors are with the QUT Centre for Robotics, School of Electrical Engineering and Robotics, Queensland University of Technology, Brisbane, QLD 4000, Australia. $^{2}$This author is affiliated with the School of Natural Sciences, Macquarie University, Sydney, NSW 2109, Australia. $^{\ast}$ Correspondence should be addressed to:
        {\tt adam.hines@qut.edu.au}}%
\thanks{This work received funding from an ARC Laureate Fellowship FL210100156 to MM and an ARC Discovery Early Career Researcher Award DE240100149 to TF. The authors acknowledge continued support from the Queensland University of Technology (QUT) through the Centre for Robotics.}
}

\usepackage{eso-pic}
\usepackage{url}
\AddToShipoutPicture*{%
     \AtTextUpperLeft{%
         \put(-3.5,10){
           \begin{minipage}{\textwidth}
              \scriptsize
              \MakeUppercase{IEEE INTERNATIONAL CONFERENCE ON ROBOTICS AND AUTOMATION (ICRA) 2026. 
              
              PREPRINT VERSION; FINAL VERSION WILL BE AVAILABLE AT \url{https://ieeexplore.ieee.org/}}
           \end{minipage}}%
     }%
}

\begin{document}
\maketitle
\bstctlcite{BSTcontrol}

\begin{abstract}
\label{sec:abstract}
Event-based localization research and datasets are a rapidly growing area of interest, with a tenfold increase in the cumulative total number of published papers on this topic over the past 10 years. Whilst the rapid expansion in the field is exciting, it brings with it an associated challenge: a growth in the variety of required code and package dependencies as well as data formats, making comparisons difficult and cumbersome for researchers to implement reliably. To address this challenge, we present Event-LAB: a new and unified framework for running several event-based localization methodologies across multiple datasets. Event-LAB is implemented using the Pixi package and dependency manager, that enables a single command-line installation and invocation for combinations of localization methods and datasets. To demonstrate the capabilities of the framework, we implement two common event-based localization pipelines: Visual Place Recognition (VPR) and Simultaneous Localization and Mapping (SLAM). We demonstrate the ability of the framework to systematically visualize and analyze the results of multiple methods and datasets, revealing key insights such as the association of parameters that control event collection counts and window sizes for frame generation to large variations in performance. The results and analysis demonstrate the importance of fairly comparing methodologies with consistent event image generation parameters. Our Event-LAB framework provides this ability for the research community, by contributing a streamlined workflow for easily setting up multiple conditions. 
\end{abstract}
\section{Introduction}
\label{sec:intro}
Autonomous robots that navigate their environment require localization systems to provide an estimate of their position. Systems used for visual localization often use Visual Place Recognition (VPR)~\cite{Masone2021} or Simultaneous Localization and Mapping (SLAM)~\cite{Placed2023} to determine a robot's position from sensor inputs, without the need for satellite-based positioning. In recent years, there has been interest in developing brain-inspired localization and mapping systems~\cite{Yu2019, Hines2025} using lessons from neuroscience, often termed \emph{neuromorphic computing}~\cite{Roy2023}. Neuromorphic localization provides the ability to develop and implement low-latency and low-energy systems for size, weight, and power (SWaP) constrained platforms, using data-rich information streams from event-based cameras~\cite{Gallego2022,Hines2025}. 

Several neuromorphic systems have been developed to perform VPR~\cite{Hines2025, Fischer2020, Lee2021, Kong2022, Fischer2022, Lee2023} and SLAM~\cite{Rebecq2017-2, Rosinol2018, Guan2023, Guo2024}, including a series of benchmark datasets~\cite{Zhu2018, Hu2020, Fischer2020, Fischer2022, Nair2024, Carmichael2024, Pan2025} to provide a community-based framework for the development and advancement of this field. Despite being early in this venture, however, there are already fragmentations in the implementation of baseline methods and datasets, which presents a challenge for the fair assessment of multiple comparisons. As new hardware and technology is developed and used by the research community, dataset formats and implementations differ, causing compatibility issues with downstream methods~\cite{Yik2025}. 

    \begin{figure}[t]
    \centering
    \vspace{4pt}
    \includegraphics[width=0.75\columnwidth]{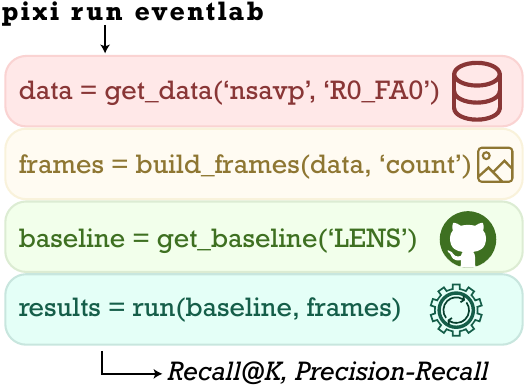}
    \vspace{-5pt}
    \caption{Overview of the Event-LAB system. Single command-line invocation for a desired baseline method and dataset triggers a series of events. Data is downloaded from an external database and setup to a standard format. Event frames (counts or reconstructions) are generated based on a configuration file consisting of multiple parameters. Baseline methods are cloned from external repositories, including any necessary model weights and checkpoints. Finally, the generated frames and baseline method is run to produce recall and precision metrics for VPR evaluation.}
    \vspace{-15pt}
    \label{fig:trainingquery}
    \end{figure}

With the advent of vastly improved code package and dependency management, we propose a unified framework to perform event-based localization and mapping evaluations across multiple baselines and datasets dubbed \emph{Event-LAB}. Using single command-line invocation, Event-LAB can be used to run multiple baseline methods with a variety of datasets in a standardized format, with evaluation metrics performed for each. This framework ensures reproducible analysis of event-based localization, streamlined workflows with minimal setup tim, and a platform for community-based contributions.

A challenge with event-based VPR is the variety of ways to define what constitutes a ‘place' in localization datasets. Methods are often developed with an event count collection or time window set as a default and fixed metric, which can translate to poor performance in certain baseline methods. In this work, we describe a generalized pipeline through Event-LAB that allows for fair, easy evaluation of baseline methods across a variety of user-defined parameters for a more complete analysis of new and emerging systems. There is a clear distinction in the performance of event-based methods when using different frame generation methods and parameters. We additionally provide an analysis of how equivalencies across different frame generation methods, event counts, and time windows can be established.

\noindent Specifically, our contributions are as follows:
\begin{enumerate}
    \item We present Event-LAB, a comprehensive unified framework to simply run event-based localization methods with multiple datasets using single command-line invocation.
    \item A thorough evaluation of available localization methods and datasets across different frame generation parameters, highlighting key advantages and disparities when comparing methods.
    \item A method to find localization equivalencies across different event count or time window collections for generating reference and query datasets.
\end{enumerate}

\noindent We have made the code publicly available at: \texttt{\small{https://github.com/EventLAB-Team/Event-LAB}}.
\section {Related Works}
\label{sec:relatedworks}
This section briefly summarizes current baseline methods for event-based localization in Section~\ref{subsec:eventbasedmethods} and datasets used to evaluate baseline methods in Section~\ref{subsec:eventbaseddata}. This is followed by a discussion of other frameworks that aim for standardizing benchmarking methods in Section \ref{subsec:benchmarks}.

\subsection{Event-based localization methods}
\label{subsec:eventbasedmethods}
For event-based localization, there are two typical means of representing the concept of a place from asynchronous event streams for baseline methods. The first is a simple event frame in which events are counted over a specified time window per pixel~\cite{Rebecq2017-2}. The second is a more complex process involving the reconstruction of images from event streams over a specified time window, using methods such as the popular E2VID~\cite{Rebecq2019, Rebecq2019-2}. Additional event frame representation methods have been developed but have not been used specifically in localization, such as Hierarchy of Time Surfaces (HOTS)~\cite{Lagorce2017}, Histograms of Averaged Time Surfaces (HATS) \cite{Sironi2018}, and Time-Ordered Recent Event (TORE) volumes~\cite{Baldwin2023}. 

For event count frames, several methods have been proposed to make conventional VPR methods compatible with the event-based representation of a place. EventVLAD~\cite{Lee2021} trains a CNN to produce reconstructed edges from event-count frames, from which features are then extracted using NetVLAD~\cite{Arandjelovic2018}. Similarly, Event-VPR~\cite{Kong2022} examined and used voxel grids of event streams for the extraction of NetVLAD features.

In contrast to these methods, Sparse-Event-VPR~\cite{Fischer2022} proposed a unique method of selecting a subset of pixels based on the variation in activity for a simple sum-of-absolute-differences (SAD) comparison with sequence matching techniques~\cite{Garg2022}. Ev-ReconNet~\cite{Lee2023} employed, for the first time, spiking neural networks to perform direct processing of event streams for edge reconstruction and NetVLAD features for energy-efficient event VPR. Most recently, Locational Encoding with Neuromorphic Systems (LENS)~\cite{Hines2025} fully integrates SNNs and a SynSense Speck neuromorphic processor for direct VPR using asynchronous event streams.

For reconstructed images, conventional and state-of-the-art VPR systems are often employed as they show impressive performance on traditional RGB images. Event reconstruction frames provide richer detail that conventional VPR methods excel with, however they come at the cost of requiring computationally expensive and time-consuming image generation. Relevant to this work, Fischer \& Milford \cite{Fischer2020} devised a strategy for using ensembles of temporal windows of event accumulation with reconstructed images showing enhanced performance for large scale traverses. VPR benchmark suites are publicly available to test a variety of conventional and state-of-the-art image-based feature extractors, such as AnyLoc~\cite{Keetha2023}, CosPlace~\cite{Berton2022}, SALAD~\cite{Izquierdo2024}, and EigenPlaces~\cite{Berton2023}. Recently, an ensembling method that used several place recognition algorithms on both frame count and reconstructed images was found to improve performance by combining the features of several algorithms \cite{Joseph2025}.

For SLAM, there are multiple methods available that perform event-based pose tracking and mapping, through both monocular and stereo depth estimation. Feature-based methods such as Ultimate-SLAM (U-SLAM)~\cite{Rosinol2018} and PL-EVIO~\cite{Guan2023} use image-like event count frames for traditional feature detection and extraction. In~\cite{Guan2022}, an event and IMU odometry (EIO) monocolar feature-based method was developed that provided real-time 6-DoF state estimation. Purely event-based monocular depth estimation was introduced by~\cite{Kim2016}, with work done by~\cite{Chiavazza2023} successfully calculating event-based optical flow asynchronously on a neuromorphic processor. 

\subsection{Event-based datasets}
\label{subsec:eventbaseddata}
Several datasets for event-based localization methods exist and range from small-scale to large-scale distances using a variety of dynamic vision sensor hardware. Small-scale datasets, such as QCR-Event-VPR~\cite{Fischer2022} and Fast-Slow-Event~\cite{Nair2024}, were generated to test specific VPR and camera setup methods. Larger-scale VPR datasets such as Brisbane-Event-VPR~\cite{Fischer2022}, Novel Sensors for Autonomous Vehicle Perception (NSAVP)~\cite{Carmichael2024}, NYC-Event-VPR~\cite{Pan2025}, and DD20 ~\cite{Hu2020} provide multi-environment, multi-condition traverses for testing of localization robustness. The seminal event-based dataset and simulator by Meuggler et al.~\cite{Mueggler2017} introduced a variety of scenarios for pose estimation, visual odometry, and SLAM. The Multi Vehicle Stereo Event Camera (MVSEC) dataset records several traverse across road and aerial environments using multiple modes of transportation~\cite{Zhu2018}. Recently, the Event Camera Rotation Dataset (ECRot) was introduced with panoramas of event-based data in motorized and hand-held sequences in up to 7K resolution~\cite{Guo2024}.

\subsection{Benchmark standardization methods}
\label{subsec:benchmarks}
To the best of our knowledge, this is the first attempt at unifying neuromorphic localization methods and datasets for event-based systems. Previous work such as NeuroBench focuses on benchmarking algorithmic and systems level operations to compare neuromorphic architectures easily~\cite{Yik2025}. For VPR-specific benchmarking, the VPR Bench method was introduced to standardize baseline method implementation across a variety of standard datasets and evaluation metrics~\cite{Zaffar2021}. A more recent extension on this work by Berton et al.~\cite{Berton2023} offers recent and current state-of-the-art systems for VPR in a simple framework. VSLAM-LAB operates on a similar framework to ours that has unified visual-SLAM methods and datasets for simple and reproducible benchmarking~\cite{Fontan2025}, and is also implemented using Pixi. 
\section {Methodology}
\label{sec:method}
Event-LAB is implemented using Pixi, allowing for multi-platform and multi-environment package management and deployment. Pixi implements project-level environments using lockfiles that capture dependency states across supported platforms, ensuring bit-for-bit reproducibility. Additionally, it resolves dependencies using Rust, making it several times faster than conventional conda environments. As Pixi makes use of multiple package ecosystems, including conda-forge and PyPI, it provides seamless integration with several robotic architectures such as ROS Noetic (via RoboStack~\cite{Fischer2022AEcosystems}).   

    \begin{table}[t]
      \caption{Implemented methods and datasets in Event-LAB.}
      \centering
      \footnotesize
      \begin{tabular}{l | l}
        \hline
        \multicolumn{2}{l}{\textbf{VPR Methods}} \\
        \hline
        LENS & Hines et al.\ 2025 \cite{Hines2025} \\
        VPR-evaluation-methods & Berton et al.\ 2022 \cite{Berton2023} \\
        Sparse-Event-VPR & Fischer \& Milford 2022 \cite{Fischer2022} \\
        EventVLAD & Lee \& Kim 2021 \cite{Lee2021} \\
        Ensemble-Event-VPR & Fischer \& Milford 2020 \cite{Fischer2020} \\
        \hline
        \multicolumn{2}{l}{\textbf{VPR Datasets}} \\
        \hline
        NSAVP & Carmichael et al.\ 2024 \cite{Carmichael2024} \\
        Fast-and-Slow & Nair et al.\ 2024 \cite{Nair2024} \\
        QCR-Event-VPR & Fischer \& Milford 2022 \cite{Fischer2022} \\
        Brisbane-Event-VPR & Fischer \& Milford 2020 \cite{Fischer2020} \\
        \hline
        \multicolumn{2}{l}{\textbf{SLAM Methods}} \\
        \hline
        Ultimate-SLAM & Rosinol et al.\ \cite{Rosinol2018} \& Rebecq et al.\ \cite{Rebecq2017} \\
        PL-EVIO & Guan et al.\ \cite{Guan2023} \\
        \hline
        \multicolumn{2}{l}{\textbf{SLAM Datasets}} \\
        \hline
        Event-camera dataset & Mueggler et al.\ \cite{Mueggler2017} \\
        \hline
      \end{tabular}
      \label{table:implemented}
        \vspace{-15pt}
    \end{table}
\subsection{Operation of Event-LAB}
\label{subsec:evop}

Pixi manages multiple package and dependency environments simply, allowing for baselines methods to be run within separate environments with a single \texttt{pixi.toml} file. This means that Event-LAB is able to work across Mac, Linux, and Windows operating systems and handle dependency differences seamlessly with packages like CUDA and PyTorch. Installing and running Event-LAB is as simple as:\medskip
\\
\begin{minted}[linenos, frame=lines, fontsize=\footnotesize, breaklines, breakanywhere]{console}
# Install pixi
curl -fsSL https://pixi.sh/install.sh | sh

# Clone the Event-LAB repository
git clone https://github.com/EventLAB-Team/Event-LAB && cd Event-LAB

# Run the evaluation
pixi run eventlab lens brisbane_event sunset2 sunset1
\end{minted}

The arguments in the command line invocation use the following standardized format:\medskip

\begin{flushleft}
\small\ttfamily
pixi run eventlab <baseline method> <dataset> <reference> <query>
\end{flushleft}
\vspace{.5\baselineskip}

Table \ref{table:implemented} highlights the currently implemented methods and datasets at the time of this work. Methods and datasets were selected on the basis of having publicly available code and data repositories. While other event-based localization methods were identified, no publicly accessible code was available to implement~\cite{Ji2023, Kong2022, Lee2023}.

\subsection{Baseline methods}
\label{subsec:basemethods}
All baseline methods (Tab.~\ref{table:implemented}) are implemented as an \texttt{EventBaseline} class that utilizes standardized evaluation metrics, individualized data formatting, and code execution blocks specific to the method.\medskip

\begin{minted}[linenos, frame=lines, fontsize=\footnotesize, breaklines, breakanywhere]{python}
class EventBaselineMethod(EventBaseline):
    def __init__(self, config):
        super().__init__()
        # Set config as class attribute
        self.config = config
        # Retrieve baseline method
        self.url = "https://github.com/<author>/<baseline_method>.git"
        if not os.path.exists(self.repo_path):
            clone_repo(self.url, destination=self.repo_path)

        # Load baseline specific configuration file
        self.baseline_config_path = './baselines/<baseline>.yaml'
        with open(self.<baseline>_config_path, 'r') as file:
            self.baseline_config = yaml.safe_load(file)
            
    def format_data(self, reference, query):
        # Dataset formatting for baseline

    def build_execute(self):
        # Builds executable command

    def run(self):
        # Runs the executable command
        
    def parse_results(self, ground_truth):
        # Performs standardized metrics

    def cleanup(self):
        # File cleanup
\end{minted}

For each baseline method (Tab.~\ref{table:implemented}), a \texttt{baseline configuration} is used to parse specific arguments or alter the implementation parameters. For example, in the VPR-evaluation-methods baseline \cite{Berton2023} it is possible to alter the model parameters such as the VPR method and backbone:\medskip

\begin{minted}[linenos,fontsize=\footnotesize,frame=lines,breaklines]{yaml}
# Configuration for the vprmethods baseline
method: mixvpr
backbone: "ResNet50"
descriptors_dimension: 512
no_labels: true
\end{minted}

\subsection{Dataset management}
\label{subsec:dataman}
Running the Event-LAB command triggers a series of checks and pre-processing tools, the first of which is the dataset management and formatting. Event-LAB standardizes datasets to the \texttt{.hdf5} file format, however it also provides conversion tools for \texttt{.bag} and \texttt{.txt} formats. This includes data formatting across different sensor types, \emph{i.e.} DAVIS and Prophesee sensors. Data is downloaded from an external source based on the dataset, reference, and query choices. A configuration file is used to determine the frame generation parameters used, parsed as a list for multiple conditions to evaluate, with settings like timestamp units determined by the user.\medskip

\begin{minted}[linenos,fontsize=\footnotesize,frame=lines,breaklines]{yaml}
frame_generator: "reconstruction"
reconstruction_model: "e2vid"
# Reconstruction time in ms
timewindows: [250, 500, 750, 1000]
# Number of events per frame for reconstruction
num_events: [25000, 50000, 75000, 100000] 
\end{minted}

Frames can also be represented by simple event counts stored as \texttt{NumPy} arrays with or without event polarity. Frames can also be generated either a fixed time window for event collection or a maximum number of events per frame.

When a dataset is called (Tab.~\ref{table:implemented}), it creates an instance of an \texttt{EventDataset} class that will check the existence of the reference and query traverses with the specified parameters, download \texttt{.hdf5} raw event files, begin the frame generation process (counts or reconstruction), and generate a pseudo ground-truth. An individual dataset configuration file is also used to define download URLs, dataset formats, availability and URL to ground-truth files, and any other additional parameters that are required.\medskip

\begin{minted}[linenos, frame=lines, fontsize=\footnotesize, breaklines, breakanywhere]{python}
class EventDataset():
    def __init__(self, config, dataset_name, sequence_name):
        super().__init__()
        # General configuration
        self.config = config 
        # Name of the dataset
        self.dataset_name = dataset_name 
        # Name of the sequence
        self.sequence_name = sequence_name 

        # Load dataset configuration
        with open(f"./datasets/{self.dataset_name}.yaml", 'r') as file:
            self.data_config = yaml.safe_load(file)
\end{minted}

A \texttt{metadata} file is generated that contains information about the event frame generation including starting dataset and sequence names, timestamp starts and durations, and event camera resolutions. The \texttt{EventDataset} class holds all the required information to run formatting of data to the .hdf5 format.

\begin{figure}[t]
    \includegraphics[width=0.9\columnwidth]{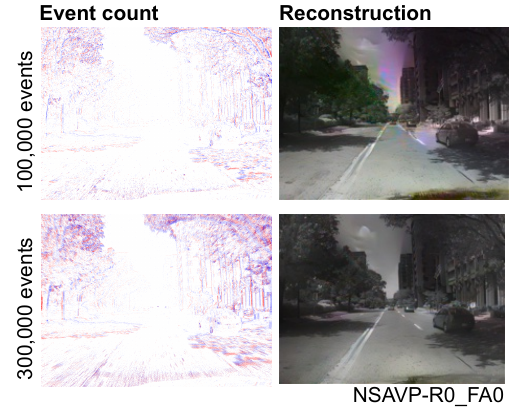}
    \caption{Example images generated from event counts (left) or reconstruction with E2VID \cite{Rebecq2019, Rebecq2019-2} (right), with 100,000 (top) and 300,000 (bottom) events per frame specified during the image generation.}
    \label{fig:imageexample}
    \vspace{-15pt}
    \end{figure}

Event images are generated either by specifying a fixed time window or the maximum number of events per frame, the latter of which will produce images of variable window length. An example of an event count and reconstructed images is shown in Fig. \ref{fig:imageexample}. Importantly, reconstruction-based methods like E2VID are recurrent networks that consider historic data in addition to newly incoming events, which in comparison to simple event counts provide richer detail that significantly improves performance in localization systems.

\subsection{Pseudo ground-truth generation}
\label{subsec:pseudogt}
Pseudo ground-truth files are auto-generated when a reference-query pair is defined in the Pixi command. The ground-truth file is a binary matrix that identifies correct reference-query matches. A ground-truth tolerance was set for each dataset based on the number of reference and query images. For small-scale, indoor datasets (QCR-Event-VPR, Fast-Slow) we set a tolerance of $\pm$ 300 msec and for the larger-scale datasets (NSAVP, Brisbane-Event-VPR), $\pm$ 10 seconds from the identified reference and query match in the pseudo ground-truth. 

Matches are identified either by global positioning system (GPS) coordinates synced to event timestamps, or customized logic based on the dataset used (odometry, start and end times). We filter matches in the reverse direction of travel, and avoid matching the beginning and the end of a traverse if they overlap. References and queries can also be filtered by time to reduce the number of images in a dataset. Tolerances and filters are fully configurable parameters. 

\begin{minted}[linenos,fontsize=\footnotesize,frame=lines,breaklines]{yaml}
# Signifying filter images every 60 seconds 
filter_time_sec: 60 
# Signifying 3 places
ground_truth_tolerance: 3 
\end{minted}
\section{Experimental Setup}
\label{sec:expsetup}
We ran all experiments, testing, and development on an Ubuntu 24.04 system running an Nvidia RTX2080 graphics processor for CUDA acceleration. To reconstruct event frames with E2VID, we alternatively generated images in parallel using a high-performance-computing (HPC) cluster running multiple NVIDIA H100 graphics processors. Our dataset reference and query pairs for VPR evaluation were the following: Brisbane-Event-VPR dataset reference: \texttt{sunset2}, query: \texttt{sunrise} \cite{Fischer2020}, NSAVP reference: \texttt{R0\_FA0}, query: \texttt{R0\_FS0} \cite{Carmichael2024}, Fast-Slow reference: \texttt{r\_high1}, query: \texttt{q\_high1} \cite{Nair2024}, and QCR-Event-VPR reference: \texttt{normal1}, query \texttt{fast2} \cite{Fischer2022}. To set up a reference and query dataset, we sampled places every 1000 msec and 100 msec for the long-scale (Brisbane-Event-VPR \& NSAVP) and short-scale (QCR-Event-VPR \& Fast-Slow) datasets, respectively. This produced on average $\approx$600-800 reference and query images per dataset. Importantly, this set of reference-query pairs is just an example and any combinations are possible for the available datasets implemented (Tab.~\ref{table:implemented}).

\subsection{Evaluation metrics and statistics}
\label{subsec:evalstats}
All baseline methods and datasets are evaluated using Recall@K and the area under the precision-recall curve PR-AUC, to evaluate the overall performance and accuracy of each system. Precision and recall are calculated as follows:

    \begin{align}
    \begin{gathered}
        Precision = \frac{TP}{TP + FP} \hskip2em
        Recall = \frac{TP}{GTP}
    \label{eq:precrecall}
    \end{gathered}
    \end{align}
    where $TP$ is the number of true positives, $FP$ is the number of false positives, and $GTP$ is the number of ground truth positives. $Recall@K$ measures the top $K$ matches of all references per query, measured at $K = [1, 5, 10]$. 

\subsection{Batching multiple experiments}
\label{subsec:batching}
Event-LAB provides a method for batching multiple baseline methods, datasets, and time windows by generating a bash \texttt{.sh} based on the \texttt{config.yaml} file:\medskip

\begin{minted}[linenos,fontsize=\small,frame=lines,breaklines]{yaml}
# Batch experiment parameters
batch_experiments:
  - dataset: "qcr_event"
    reference: "normal1"
    queries: ["normal2", "normal3", "fast2", "slow1"]
    num_events: [25000, 50000, 100000]
    frame_generator: "frames"
    frame_accumulator: "eventcount"
    baselines: ["sparse_event", "lens", "eventvlad"]
\end{minted}

In this configuration, two main experiments will run using event-frame counts and frame reconstructions, with different baseline methods and datasets. Users can build a batch experiment of their choosing, and simply run the following to generate and execute the bash script: \texttt{pixi run batch}.

The bash script modifies the configuration file before execution of each baseline method with a dataset at the specified time window. The output of each experiment is stored to a combined spreadsheet that accumulates individual Recall@K and PR-AUC curves, as well as summary statistics of each method and baseline across frame generation parameters.

In this work, for any event reconstructed images we only used the E2VID algorithm with the E2VID checkpoint \cite{Rebecq2019, Rebecq2019-2}. All event count frames generated consider event polarity, however for baseline method formatting these are summed into a single channel. In addition to setting frames with time windows, Event-LAB allows users to match the number of events per frame from references to queries in datasets that have significant velocity differences, such as in QCR-Event-VPR \cite{Fischer2022}.
\section{Results}
\label{sec:results}

\subsection{Summary results with Event-LAB}
\label{subsec:summaryresult}
\begin{figure}[t]
    \centering
    \includegraphics[width=0.9\columnwidth]{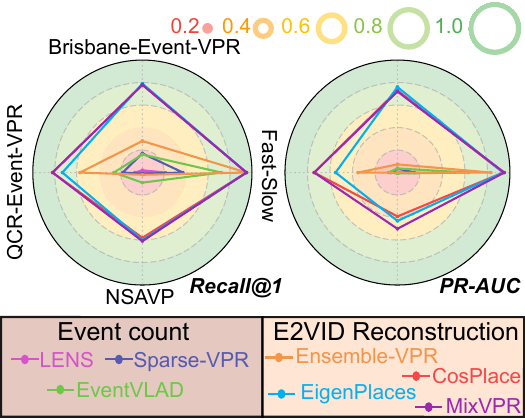}
    \vspace{-5pt}
    \caption{Summary results from Event-LAB comparing across multiple baseline methods and datasets. All event frames were created using a fixed number of events per frame: 25,000 for QCR-Event-VPR, Brisbane-Event-VPR, and Fast-Slow, and 100,000 for NSAVP. Event counts were selected based on the camera used to collect them, a DAVIS346 and DVXplorer respectively. These event counts approximated to on average 33 msec per frame. Baseline methods that used reconstructed images performed best overall (CosPlace~\cite{Berton2022}, EigenPlaces~\cite{Berton2023}, MixVPR~\cite{Ali2023}).}
    \label{fig:summaryresults}
    \vspace{-15pt}
    \end{figure}

To visually compare performance of different baseline methods and datasets, we ran Event-LAB across all currently implemented methods with all datasets. A summary of the results is shown in Fig.~\ref{fig:summaryresults}. The Recall@1 and PR-AUC results highlight a diversity of outcomes for baseline methods and datasets, owing to differences in data collection, method development, and quality of reconstructed images. 

Overall, the baseline methods that utilized reconstructed images performed best overall for Recall@1 and PR-AUC across the datasets tested~\cite{Ali2023, Berton2022, Berton2023}. Methods such as LENS and Sparse-Event-VPR did not perform as well at this smaller time window ($\approx$33 msec) as they were both developed for longer event collection times of 1000 msec~\cite{Fischer2022, Hines2025}. For Ensemble-Event-VPR, the original implementation used a curated reference and query set that focused on finding unique features to assist in matching~\cite{Fischer2020}. In Event-LAB, the focus is the ability to construct event based frames simply and easily across any time or event count domain desired. Results from Fig.~\ref{fig:summaryresults} can easily be obtained with a single batch script that will obtain raw data, generate images, and run baselines with a single terminal command. 

\subsection{Fixed time window performance}
\label{subsec:fixedtime}
Creating event frames by event count generates images of varied time window length. To evaluate the performance effects of different times across fixed time window frame generation, we used the Fast-Slow dataset~\cite{Nair2024} with both frame count and reconstruction methods to evaluate Recall@1 and PR-AUC performance. The results are summarized in Table \ref{table:fixedtime}. For the fixed time analysis, we did not subsample by time as the entire reference and query set was used to establish performance gains and losses through using different time windows.

    \begin{table}[t]
      \setlength{\tabcolsep}{1.75pt} %
      \caption{Fixed time-window performance (Recall@1 and PR-AUC) across baselines from the Fast-Slow dataset with r\_med1 as the reference and q\_med1 as the query. \textbf{Bold} is the best baseline method for both frame count and reconstructed baselines. All times are in msec.}
      \centering
      \footnotesize
      \begin{tabular}{l | c c c c c c}
        \hline
        \textbf{Baseline method} & \textbf{33} & \textbf{66} & \textbf{120} & \textbf{250} & \boldmath$\mu$ & \textbf{PR-AUC} \\
        \hline\hline
        LENS~\cite{Hines2025}           & 0.05 & 0.16 & 0.53 & 0.77 & 0.38 & 0.29 \\
        Sparse-Event~\cite{Fischer2022} & 0.07 & 0.20 & 0.52 & 0.83 & 0.56 & 0.39 \\
        EventVLAD~\cite{Lee2021}        & \textbf{0.72} & \textbf{0.84} & \textbf{0.87} & \textbf{0.95} & \textbf{0.84} & \textbf{0.76} \\
        \hline
        Ensemble-Event~\cite{Fischer2020} & \textbf{0.99} & \textbf{1.0} & \textbf{1.0} & \textbf{1.0} & \textbf{0.99} & \textbf{0.99} \\
        CosPlace~\cite{Berton2022}        & 0.88 & 0.92 & 0.95 & \textbf{1.0} & 0.94 & 0.84 \\
        EigenPlaces~\cite{Berton2023}     & 0.89 & 0.92 & 0.95 & \textbf{1.0} & 0.94 & 0.88 \\
        MixVPR~\cite{Ali2023}             & 0.86 & 0.92 & 0.97 & \textbf{1.0} & 0.94 & 0.93 \\
        \hline
      \end{tabular}
      \label{table:fixedtime}
    \end{table}

In fixed time windows, the 33 msec condition led to results similar to those obtained in Fig.~\ref{fig:summaryresults}. For baseline methods such as LENS and Sparse-Event-VPR, performance significantly improves with increasing time window lengths, since these methods were developed and excel at processing longer event-collection durations.~\cite{Fischer2022, Hines2025}.

Overall, the best performing methods were still ones that used traditional and state-of-the-art VPR feature extractors. We note however that Ensemble-Event-VPR works best in the reconstructed methods as it projects multiple windows onto the largest time, discarding all frames in between~\cite{Fischer2020}, leading to higher performance at lower msec time windows.

Across both event count and fixed time window evaluations, image reconstructions show the clearest advantage by being able to extract higher quality features than those available from simple event count frames. 

\subsection{Evaluating matching across collection windows}
\label{subsec:evalwindows}
\begin{figure}[t]
    \includegraphics[width=\columnwidth]{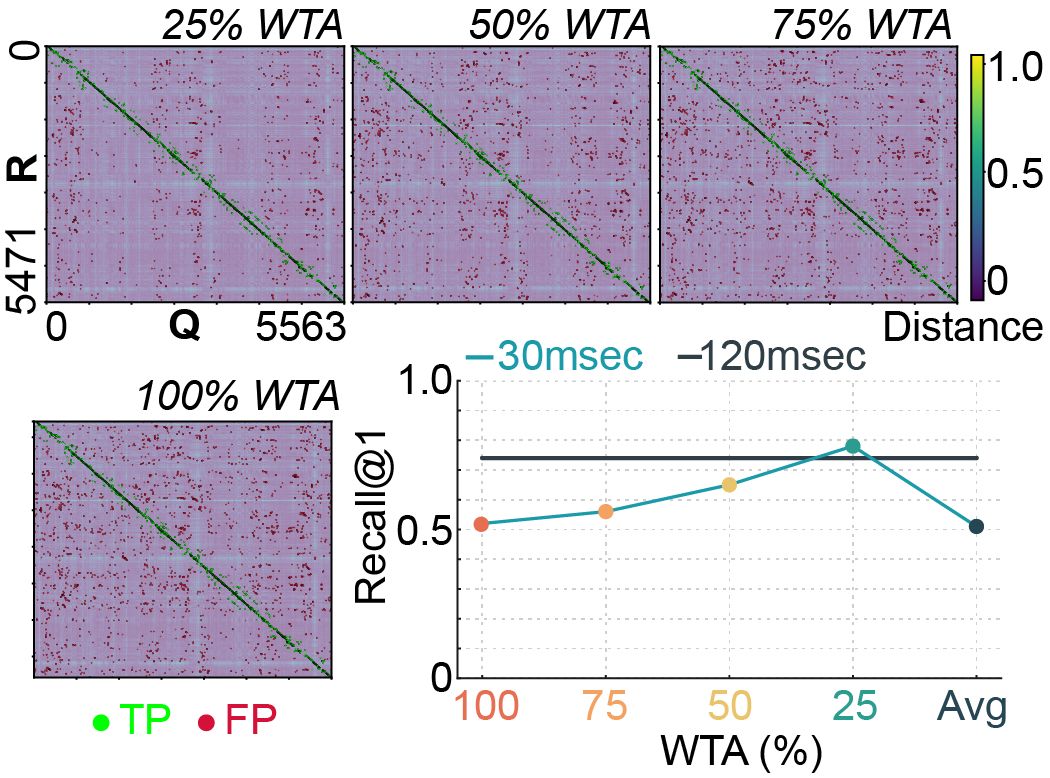}
    \vspace{-15pt}
    \caption{Evaluation of matches from a 30 msec to 120 msec images using a winner-takes-all (WTA) approach to account for varied collections lengths. Distance matrices for the $0\%$, $25\%$, $50\%$, and $75\%$ WTA condition have the true positives (TP) and false positives (FP) overlaid. The less strict $25\%$ condition allows for a significant reduction in FP, whereas those in the $75\%$ condition does not improve much. All WTA methods improved the Recall@1 over the $100\%$ baseline, however averaging across references and queries in the 30 msec had no effect. The baseline method used was EventVLAD~\cite{Lee2021} and the dataset was QCR-Event-VPR~\cite{Fischer2022}, with \texttt{normal1} and \texttt{normal2} as the reference/query pair.}
    \label{fig:evalwindows}
    \vspace{-15pt}
    \end{figure}

As higher time windows or event counts can lead to better performance, the question of whether equivalency from lower windows or event counts can be found to improve Recall@1 was raised. To do so, a winner-takes-all (WTA) scheme was devised such that if a percentage of the matches within a time window bin are correct and the longer-window single match for that same period is also correct, then we mark the whole period as a true positive. For example, if we consider a place with a 120 msec time window and $4 \times 30$ msec places in the same bin, if $50\%$ of the 30 msec matches were correct and the 120 msec place was also correct, we consider all $4 \times 30$ msec places as correct. We additionally evaluated whether simply averaging over smaller time windows to match larger ones could achieve a similar outcome.  Fig.~\ref{fig:evalwindows} highlights this and shows the changes in performance using EventVLAD~\cite{Lee2021} and QCR-Event-VPR~\cite{Fischer2022} as an example. All WTA implementations improved the Recall@1 performance, with the more relaxed $25\%$ WTA condition exceeding performance of the 120 msec condition (Fig.~\ref{fig:evalwindows}. Averaging across 30 msec time windows failed to improve Recall@1 at all.

    \begin{figure*}[t]
    \centering
    \includegraphics[width=\textwidth]{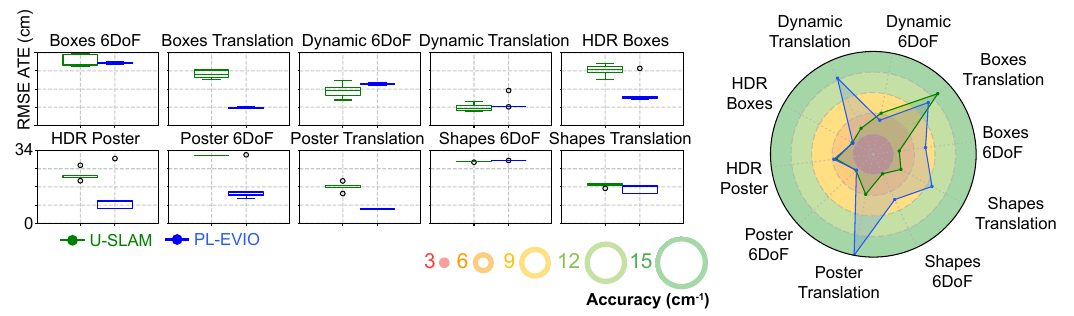}
    \vspace{-15pt}
    \caption{Metrics for the SLAM baseline methods, U-SLAM~\cite{Rosinol2018} and PL-EVIO~\cite{Guan2023}, using 5 different scenes and motions from the Event-Camera Dataset and Simulator~\cite{Mueggler2017}. The box plots (left) show the RMSE-ATE in $cm$ across 5 individual trials. The radar plot (right) shows the Accuracy in $cm^{-1}$, which is the inverse of RMSE-ATE.}
    \label{fig:slammetrics}
    \end{figure*}

    \begin{figure}[t]
    \includegraphics[width=\columnwidth]{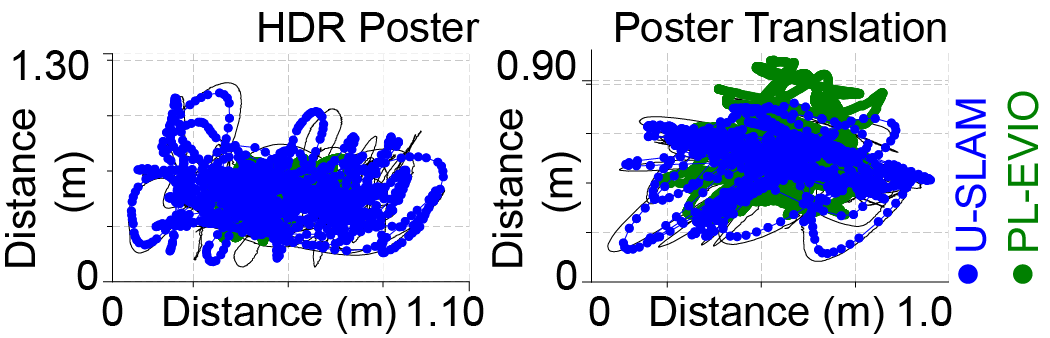}
    \vspace{-15pt}
    \caption{Example SLAM trajectories from the Poster scene from the Event-Camera Dataset and Simulator~\cite{Mueggler2017}. Black is the ground truth trajectory.}
    \label{fig:slamtraj}
    \end{figure}

Quantitatively, the equivalent is also true for generating frames through event counts, with a $14\%$ increase in Recall@1 applying a $50\%$ WTA rule to event frames generated with 25,000 events when compared to frames generated with 100,000 events. Similar to using ground truth tolerances, a WTA approach for event based frames could be considered a means of easing overly strict matching criteria.

\subsection{Event-based Simultaneous Localization and Mapping}
\label{subsec:slameval}
In order to evaluate SLAM baseline methods for Event-LAB, we measured the Root Mean Squared Error Absolute Trajectory Error (RMSE-ATE) across various motions and scenes from the Event-Camera Dataset and Simulator~\cite{Mueggler2017}. The estimated and ground-truth trajectories were aligned with a 6-DoF transformation in $\mathbf{SE}(3)$, using the entire trajectory. Fig.~\ref{fig:slammetrics} summarizes the main results, showing the RMSE-ATE and accuracy across five different scenes. PL-EVIO~\cite{Guan2023} was found to perform the best with the lowest degree of trajectory error across all scenes. Fig.~\ref{fig:slamtraj} shows example trajectories from the best performing scene (Poster).

Both U-SLAM and PL-EVIO use event frames to perform feature extraction and tracking to estimate trajectories, additionally integrating traditional RGB frame based images and inertial measurement unit (IMU) data for enhanced performance. Both SLAM methods ran in real time and covered $100\%$ of the recording without any failures. The promising initial results with the simple Event-LAB implementation provide the platform for integration with additional datasets and methods in the future.

\subsection{Setup time and metrics}
\label{subsec:setup}
An important component of Event-LAB is the ease of setup and the rapid formatting and output of recall and precision data. To evaluate benchmarking capability and times, we set up three independent experiments and measured the time taken to 1) obtain the data files and baseline method project repositories, 2) generate the event frames, and 3) run and obtain the results, including the installation of Pixi and environment building. The details and results are summarized in Table~\ref{table:setup}. Simple VPR methods and small-scale datasets like Sparse-Event-VPR~\cite{Fischer2022} and the Fast-Slow dataset~\cite{Nair2024} are very quick to download and run, whereas larger-scale and computationally expensive processes like EventVLAD~\cite{Lee2021} and NSAVP~\cite{Carmichael2024} take more time to generate frames and run the baseline method.

\begin{table}[t]
  \caption{Setup time for Event-LAB for various baselines and datasets}
  \centering
  \setlength{\tabcolsep}{1.75pt} %
  \begin{adjustbox}{width=\columnwidth}
  {\scriptsize
  \begin{tabular}{c | c c c c c c}
    \hline
    \textbf{Method} & \textbf{Dataset} & \textbf{Query} & \textbf{Ref.} & \textbf{Type} & \textbf{Events} & \textbf{Time (mm:ss)} \\
    \hline\hline
    Sparse~\cite{Fischer2022} & QCR~\cite{Fischer2022} & fast1 & normal1 & Count & 25\,000 & 01:01 \\
    MixVPR~\cite{Ali2023} & Fast-Slow~\cite{Nair2024} & q\_low1 & r\_low1 & Recons. & 25\,000 & 06:35 \\
    EventVLAD~\cite{Lee2021} & NSAVP~\cite{Carmichael2024} & R0\_RN0 & R1\_DA0 & Count & 250\,000 & 47:42 \\
    \hline
  \end{tabular}
  }
  \end{adjustbox}
  \label{table:setup}
  \vspace{-15pt}
\end{table}

We note that with high performance computing (HPC) clusters, this process becomes entirely parallelizable, running multiple baselines and event frame generation simultaneously. The main bottleneck in the Event-LAB process is the event frame generation, particularly for image reconstruction, and in some cases downloading data files from host servers. Once frames have been generated and stored, the runtime for experiments significantly decreases.
\section{Discussion and Conclusions}
\label{sec:discuss}

We present Event-LAB, a simple and reproducible means to run and contribute event-based localization methods for the greater research community. The goal was to provide a community-based tool that continues to have new methods and datasets added and to create a platform allowing for simple additions. Event-LAB will expand to additional localization and navigational methods such optical flow and classification tasks such as object detection and recognition. Here, we focused only on accuracy metrics however we do note that method latencies and power consumption are also important considerations in event-based systems, but was outside the scope for investigation and discussion.

In this work, we identify several potential future directions. We propose a new metric for averaging the varied performance across different event frame collection parameters through a winner-takes-all strategy. Future work will focus on developing a robust matching analysis pipeline to assist in reducing the computational overhead of event-based localization methods by minimizing the generation of thousands of low-event-count or short-time-window places. An additional focus in this regard is to develop the ability for asynchronous localization streaming, without the dependence on the creation of event frames for evaluation. We will additionally focus on including SNN-based localization pipelines for asynchronous processing, streaming directly from event files. This future direction directly addresses a common issue in the literature for event-based methodologies that rely on conventional representations of event streams for localization tasks. Finally, we propose that Event-LAB could in the future be used to fuse features and output from multiple localization methods for accurate, event-based navigation pipelines.
\vspace{-5pt}

\bibliography{Bibliography}
\end{document}